\documentclass[conference]{IEEEtran}
\IEEEoverridecommandlockouts
\usepackage{cite}
\usepackage{amsmath,amssymb,amsfonts}
\usepackage{algorithmic}
\usepackage{graphicx}
\usepackage{bm}
\usepackage{textcomp}
\usepackage{xcolor}
\usepackage{subfigure}

\def\BibTeX{{\rm B\kern-.05em{\sc i\kern-.025em b}\kern-.08em
    T\kern-.1667em\lower.7ex\hbox{E}\kern-.125emX}}
\begin{document}
\title{A Class of Dual-Frame Passively-Tilting Fully-Actuated Hexacopter}

\author{
\IEEEauthorblockN{Jiajun Liu}
\IEEEauthorblockA{\textit{School of Astronautics} \\
\textit{Harbin Institute of Technology}\\
Harbin, China \\
jiajunliu@stu.hit.edu.cn}
\and
\IEEEauthorblockN{Yimin Zhu}
\IEEEauthorblockA{\textit{School of Astronautics} \\
\textit{Harbin Institute of Technology}\\
Harbin, China \\
ymzhu@hit.edu.cn}
\and
\IEEEauthorblockN{Xiaorui Liu}
\IEEEauthorblockA{\textit{School of Astronautics} \\
\textit{Harbin Institute of Technology}\\
Harbin, China \\
xiaoruiliu@stu.hit.edu.cn}
\\
\and
\IEEEauthorblockN{Mingye Cao}
\IEEEauthorblockA{\textit{School of Astronautics} \\
\textit{Harbin Institute of Technology}\\
Harbin, China \\
mycao@stu.hit.edu.cn}
\and
\IEEEauthorblockN{Mingchao Li}
\IEEEauthorblockA{\textit{School of Astronautics} \\
\textit{Harbin Institute of Technology}\\
Harbin, China \\
mingchaoli@stu.hit.edu.cn}
\and
\IEEEauthorblockN{Lixian Zhang, \textit{Fellow}, IEEE}
\IEEEauthorblockA{\textit{School of Astronautics} \\
\textit{Harbin Institute of Technology}\\
Harbin, China \\
lixianzhang@hit.edu.cn}
}

\maketitle

\begin{abstract}
This paper proposed a novel fully-actuated hexacopter. It features a dual-frame passive tilting structure and achieves independent control of translational motion and attitude with minimal actuators. Compared to previous fully-actuated UAVs, it liminates internal force cancellation, resulting in higher flight efficiency and endurance under equivalent payload conditions. Based on the dynamic model of fully-actuated hexacopter, a full-actuation controller is designed to achieve efficient and stable control. Finally, simulation is conducted, validating the superior fully-actuated motion capability of fully-actuated hexacopter and the effectiveness of the proposed control strategy.
\end{abstract}

\begin{IEEEkeywords}
Fully-actuated hexacopter, mechanics, modeling and control.
\end{IEEEkeywords}

\section{Introduction}
In recent years, rotorcraft unmanned aerial vehicles (UAVs) have witnessed significant growth in scientific, military, and commercial applications\cite{b1,b2,b3,b4,b5,b6}. However, conventional rotorcraft UAVs rely on attitude adjustments to achieve positional movement in any direction, which severely constrains flight stability and limits application potential. To address these limitations, fully-actuated UAVs capable of decoupling attitude and position through full six-degree-of-freedom control have emerged as a promising alternative. Such platforms can perform physical interactions with the environment at arbitrary positions while maintaining a specific orientation, offering enhanced stability and maneuverability compared to traditional UAVs\cite{b7,b8,b9,b10,b11,b12}.

Based on research in recent years on fully-actuated UAVs, they can primarily be categorized into two types: fixed-tilt and dynamic-tilt. The fixed-tilt type features a simple and lightweight structure, ease of maintenance, high rigidity, and relatively low mechanical vibration introduced during flight.
\begin{figure}[tbp]
	\centering
	\includegraphics[width=0.47\textwidth]{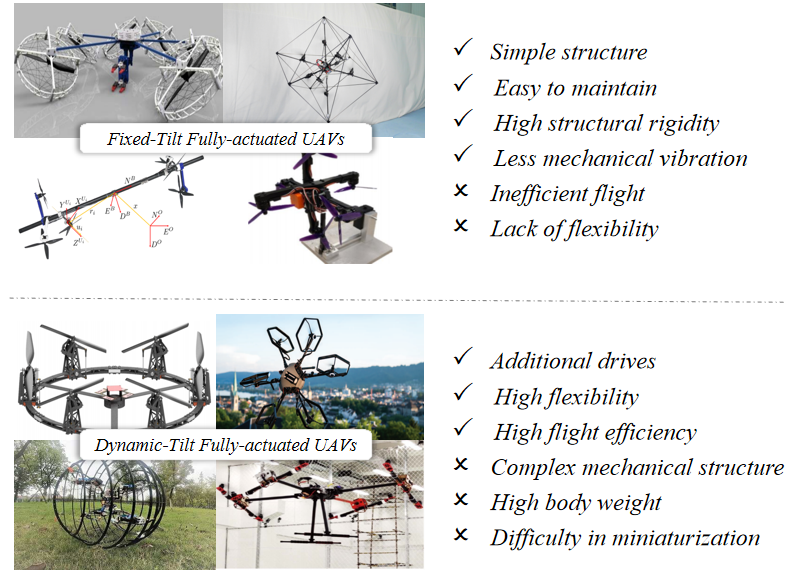}
	\caption{Summary of the fully-actuated UAVs, which are roughly divided into two categories: fixed-tilt ones and dynamic-tilt ones, cf.}
	\label{intro}
		\vspace{-2.0em}
\end{figure}
In 2018, Staub et al. at the University of Toulouse developed an asymmetric fixed-tilt hexacopter with enhanced lateral disturbance rejection\cite{b13}, yet its fixed geometry induced considerable thrust cancellation. In response to this, Brescianini et al. at ETH Zurich introduced a hexahedron-based omnidirectional octocopter with optimized thrust vectors\cite{b14}, though its complex structure increased weight and power consumption. On another front, bidirectional rotor architectures have gradually gained attention.
In 2016, Park and colleagues at Seoul National University in South Korea developed the asymmetrically distributed ODAR-6 hexacopter\cite{b15}. By utilizing bidirectional propeller thrust, the vehicle achieved consistent hover performance in all directions. In 2022, a team led by Hyungyu at the University of Illinois in the United States introduced a geometrically symmetric, bidirectional-rotor hexacopter\cite{b16}. They designed a geometric controller for this platform, successfully validating its stability and trajectory tracking capability during omnidirectional flight.
Nevertheless, bidirectional rotor systems still face issues such as response delay and oscillation. The complex transmission mechanisms also contribute to increased system weight, potential failure risks, and higher control algorithm complexity. Overall, while fixed-tilting structures offer a degree of omnidirectional flight capability, their inherent thrust cancellation issue considerably reduces flight efficiency.

The dynamic-tilt type employs additional drive mechanisms to actively tilt rotors, thereby expanding its operational range, avoiding thrust cancellation, and improving flight efficiency. 
In 2016, Ryll et al. from the University of Toulouse proposed FAST-Hex, a tilt-rotor UAV with single tilting mechanism enabling mode switching between underactuated and fully-actuated states\cite{b17}, though rotor coupling caused whole-machine resonance. In response to this, in 2018, Kamel et al. at ETH Zurich introduced Voliro\cite{b18}, a constraint-free omnidirectional tilt-rotor hexacopter for improved agility. In 2019, Li Binbin's team at Southwest Jiaotong University developed BioTetra, a tetrahedral UAV with two-degree-of-freedom rotor tilting for optimized lift\cite{b19}, but its independent servo mechanisms increased complexity and weight. In 2024, Qin et al. from Sun Yat-sen University proposed a tilt-rotor octocopter with passive hinges to reduce servo reliance\cite{b20}, though it added weight from hinge and arm structures.
In summary, the dynamic-tilt UAV achieves improved hover efficiency and omnidirectional flight capability at the expense of additional mechanical complexity and increased airframe weight. This structure is difficult to miniaturize, and when scaled up, it is hard to ensure longevity and stability.

To address the aforementioned shortcomings of fully-actuated UAVs, we propose a highly energy-efficient and structurally concise fully-actuated hexacopter. The main contributions of this work are as follows:

\textit{(i)}This paper presents a novel hexacopter with a dual-frame passive tilting mechanism that achieves full actuation using minimal actuators. The design is structurally simple, robust, and maintainable. Unlike conventional fully actuated UAVs, it eliminates internal thrust cancellation, directing all rotor thrust beyond gravity solely to flight control. This optimizes the efficiency-actuator trade-off, leading to significantly extended endurance.

\textit{(ii)}Regarding this innovative configuration, a dynamic model of the hexacopter was established, and on this basis, a novel hierarchical control system architecture for the fully-actuated hexacopter was designed.

\textit{(iii)} The flight performance and full actuation characteristics of the hexacopter were verified through simulation studies conducted.

The remainder of this paper is organized as follows: Section II describes the mechanical design details of fully-actuated hexacopter. Section III presents the dynamic modeling of hexacopter. Section IV elaborates on the full-actuation controller design for the system. Section V provides and interprets the experimental results of hexacopter. Finally, Section VI concludes the paper.

\section{Mechanical Design}

\begin{figure}[htbp]
	\centering
	\includegraphics[width=0.4\textwidth]{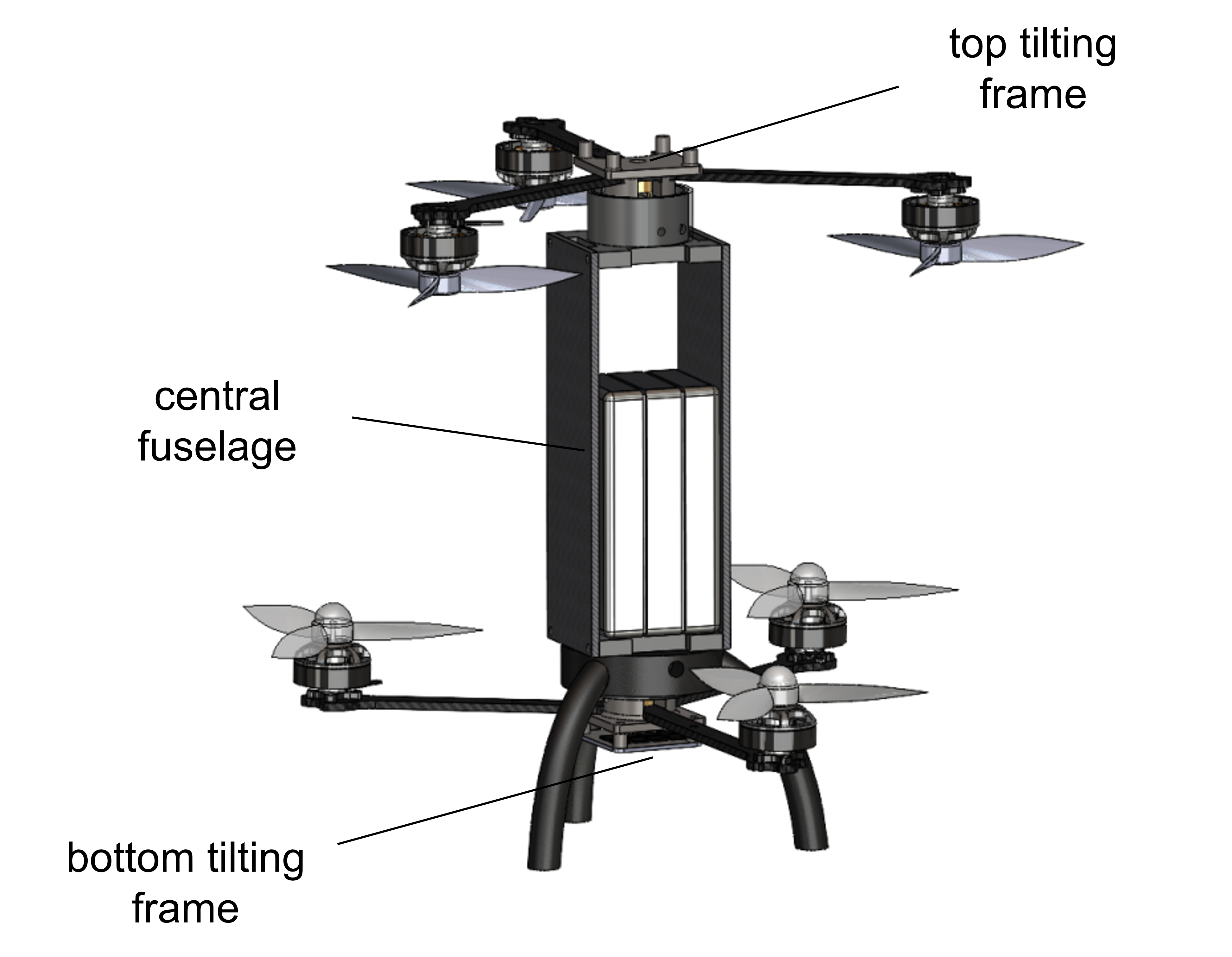}
	\caption{The mechanical design details of the fully-actuated hexacopter.}
	\label{coordinate_frames}
		\vspace{0em}
\end{figure}

As shown in Fig \ref{coordinate_frames}, the fully-actuated hexacopter employs a dual-frame passively tiltable structure. The central fuselage serves as the primary load-bearing structure carrying major payloads, while the passive tilting frames (one on the top and one on the bottom) are connected to the fuselage via universal joint.The three rotors on each frame rotate in the same direction, providing thrust while also enabling pitch and roll motions of the frame. As a result, the fuselage receives two independent resultant thrust vectors from the top and bottom frames. Furthermore, the universal joint used in this design can transmit torque along the z-axis. The rotors on the top and bottom frames rotate in opposite directions, thereby applying a counter-torque pair about the z-axis to control yaw. Thus, all six degrees of freedom corresponding to the position and attitude of the hexacopter can be independently controlled, achieving full actuation. Additionally, unlike fully-actuated UAVs with fixed tilt angles, in this design, during hover, all six rotors generate thrust purely vertically to counteract gravity, maximizing thrust utilization efficiency.

However, the implemented universal joint is non-ideal; it exhibits uniform degrees of freedom only within a specific angular range. To maintain operation within this range, mechanical limit sleeves are incorporated to restrict the tilt angles of both the upper and lower frames to within 20°. Consequently, the reference angles in the controller design are likewise constrained within this 20° bound, thereby ensuring consistency with the dynamic model assumptions and improving modeling accuracy. The selection of this angular threshold also achieves a balance between structural safety and actuation performance. Most of the payload in this design is placed on the fuselage. This approach minimizes the frames' moment of inertia to the greatest extent, ensuring effective angular response capabilities. A top view of this design is shown in Fig \ref{exp3}. The six propellers do not overlap and are arranged at 60-degree intervals, ensuring no overlap or interference between the airflow generated by each motor, thereby further reducing thrust losses.

\section{Dynamic Modeling}

The body coordinate system, denoted as $\bm{B}$, is rigidly attached to the hexacopter airframe. Its origin coincides with the center of mass of the fully-actuated hexacopter. The $\bm{B_x}$ axis is aligned with the forward direction of the vehicle, the $\bm{B_y}$ axis lies in the bodily plane and is perpendicular to the $\bm{B_x}$ axis, pointing to the right, and the $\bm{B_z}$ axis is determined by the right-hand rule (perpendicular to the bodily plane, pointing upward). The position of frame $\bm{P}$ with respect to the inertial frame $\bm{I}$ is denoted by $\bm{p}= \begin{bmatrix} x_m & y_m & z_m \end{bmatrix}^T$.

\begin{figure}[htbp]
	\centering
	\includegraphics[width=0.3\textwidth]{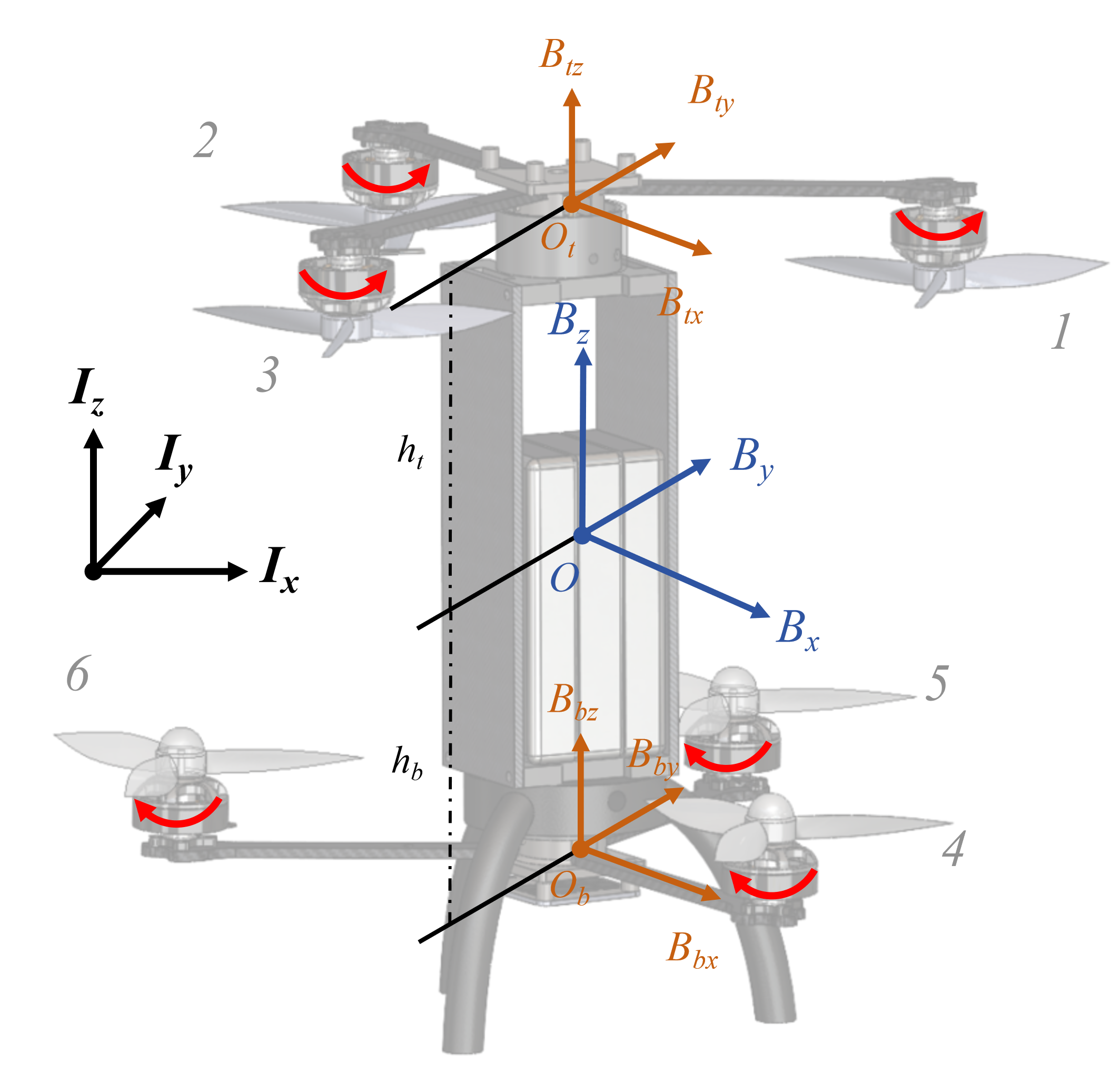}
	\caption{The definition of the coordinate systems of the fully-actuated hexacopter, comprising the world inertial coordinate system $\bm{I}$, the body coordinate system $\bm{B}$, the top-body coordinate systems $\bm{B_t}$, and the bottom-body coordinate systems $\bm{B_b}$.}
	\label{coordinate_system}
		\vspace{0em}
\end{figure}

Furthermore, subordinate body coordinate systems are established: $\bm{B_t}$ denotes the top-body coordinate system, while $\bm{B_b}$ represents the bottom-body coordinate system. Their origins, ${O_t}$ and ${O_b}$, are located at the geometric centers of the top and bottom frames, respectively. The coordinate axes of both $\bm{B_t}$ and $\bm{B_b}$ are parallel to those of the body coordinate system $\bm{B}$. Their position vectors with respect to $\bm{B}$ are denoted by $\bm{p_t^B}=[ \begin{array} {ccc}0 & 0 & h_t \end{array}]^T$ and $\bm{p_b^B}=[ \begin{array} {ccc}0 & 0 & -h_b \end{array}]^T$, respectively. ${h_t}$ and ${h_b}$ represent the vertical distances from the geometric centers of the top and bottom frames to the center of mass of the hexacopter.

Additionally, $\bm{P_t}$ and $\bm{P_b}$ are used to denote the top-frame and bottom-frame coordinate systems, respectively. The origins of these two coordinate systems coincide with the geometric centers of the top and bottom frames. Their coordinate axes rotate with the motion of the universal joint mechanism. Specifically, the $\bm{P_{tx}-P_{ty}}$ plane is defined as the plane of rotation for the top frame, while the $\bm{P_{tx}-P_{ty}}$ plane is defined as the plane of rotation for the bottom frame.

\begin{figure}[htbp]
	\centering
	\subfigbottomskip=2pt 
	\subfigcapskip=-3pt 
	\subfigure[Top view]{
		\label{Circle experiment}
		\includegraphics[width=0.3\linewidth]{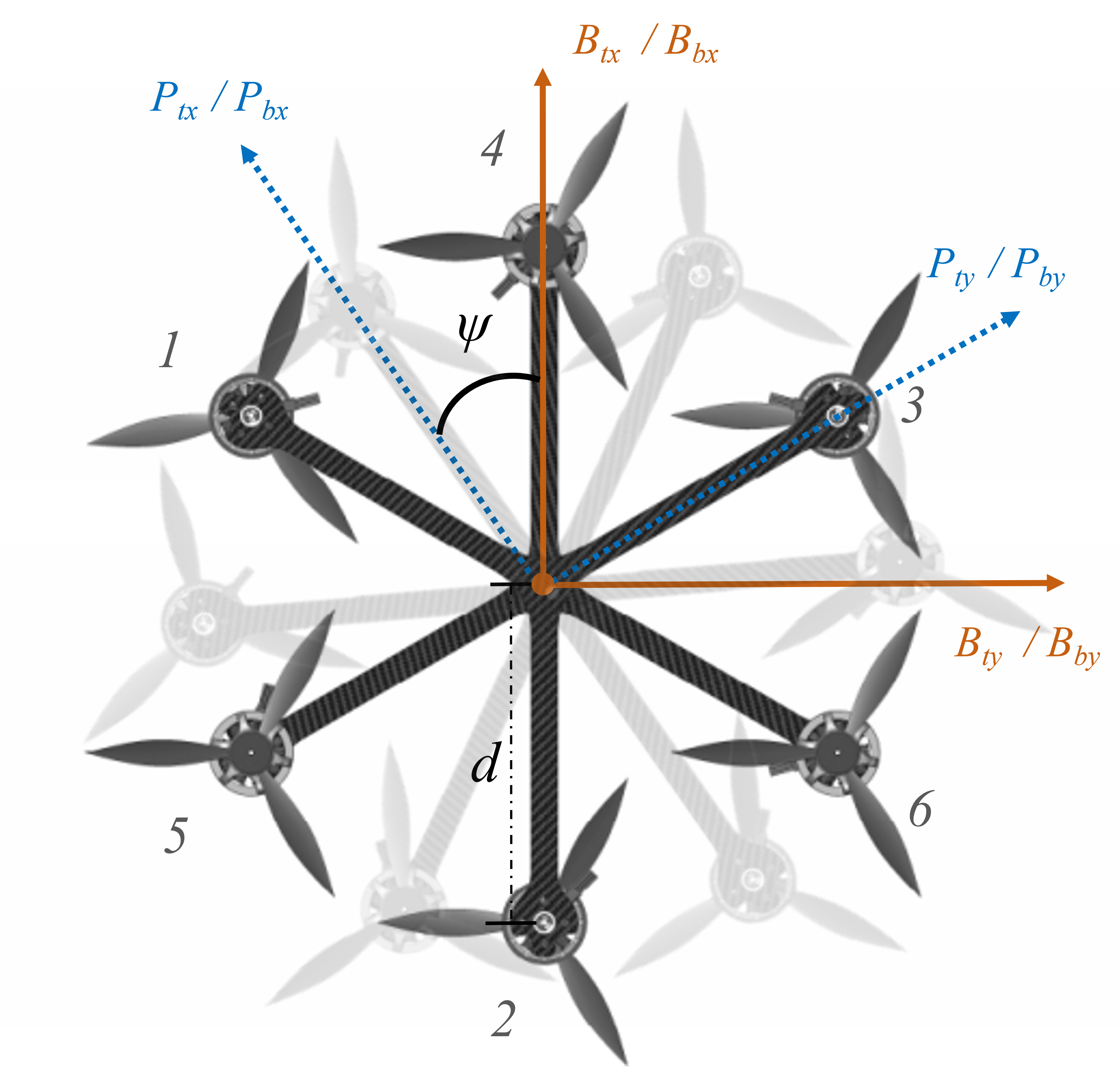}} 
	\subfigure[Right view]{
		\label{Letters experiment}
		\includegraphics[width=0.3\linewidth]{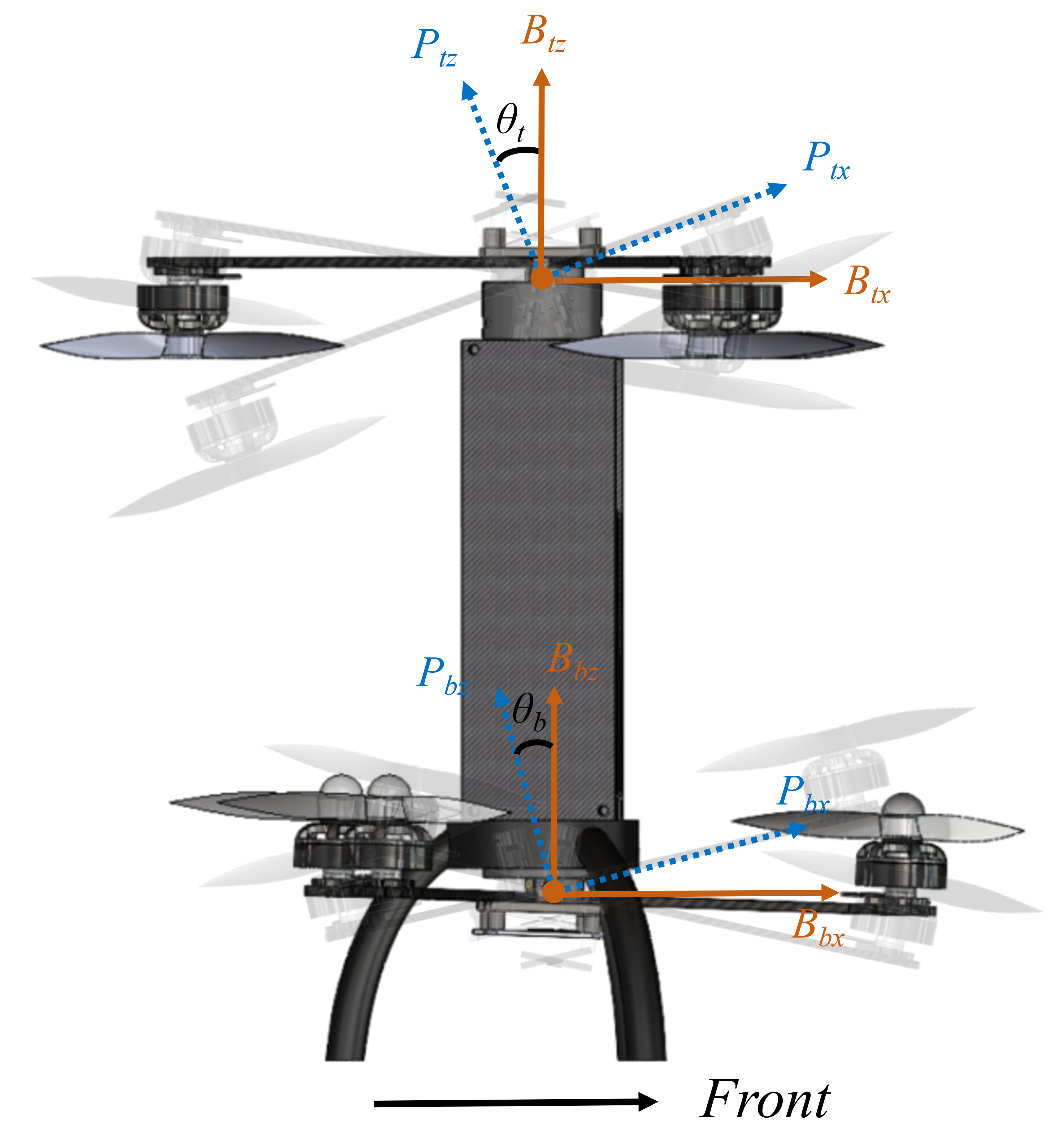}}
	\subfigure[Front view]{
		\label{Other experiment}
		\includegraphics[width=0.3\linewidth]{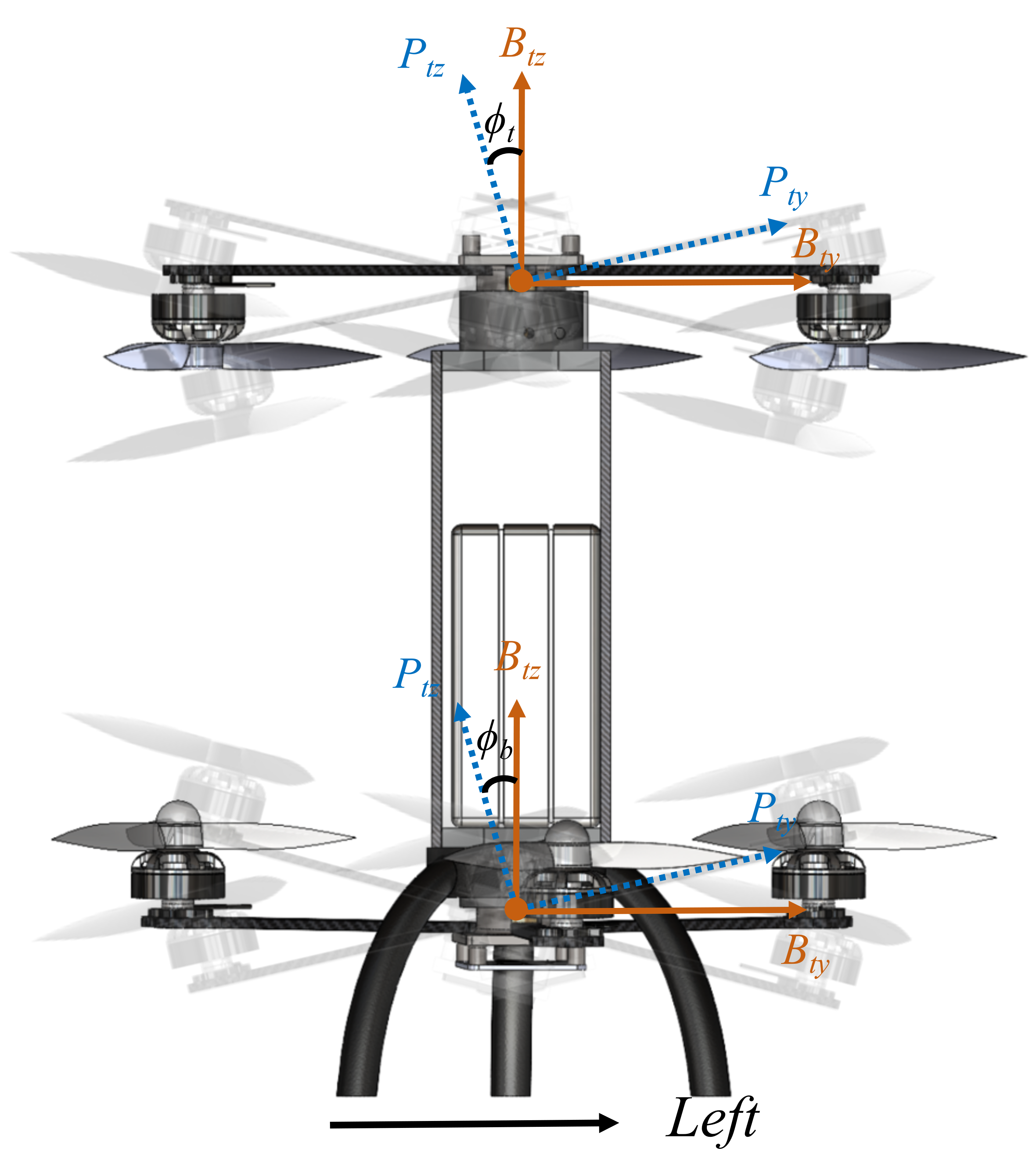}}
	\caption{The definition of the coordinate systems of the fully-actuated hexacopter, comprising the top-frame coordinate systems $\bm{P_t}$ and bottom-frame coordinate systems $\bm{P_b}$.}
	\label{exp3}
	\vspace{-1.em}
\end{figure}

Taking the dynamic modeling of the top frame as an example, its motion is described using Euler angles (following the $\bm{Z}$-$\bm{Y}$-$\bm{X}$ rotation sequence convention commonly adopted in aircraft attitude representation). Due to the mechanical constraints imposed by the universal shaft, the frame possesses no rotational degree of freedom about the $\bm{P_{tz}}$ axis (yaw direction). The tilt angles of the rotation planes of the top frame are defined as $\theta{t}$ and $\phi{t}$, as illustrated in the Fig \ref{exp3}. These angles $\theta{t}$ and $\phi{t}$ have a clear geometric interpretation: they represent the rotation angles about the $\bm{Y}$-axis and $\bm{X}$-axis of the top-body coordinate system $\bm{B_{t}}$, applied sequentially. The rotation matrix from the top-frame coordinate system $\bm{P_{t}}$ to the top-body coordinate system $\bm{B_{t}}$ is denoted as $\bm{{}^{B_t}R_{P_t}}$, whose specific form is given by:

\begin{equation}
	\begin{split}
		&\bm{R_{P_{t}}}  = \bm{R_{x}(\phi_{t})}\bm{R_{y}(\theta_{t})} \\ 
		& = \begin{bmatrix}
			\cos(\theta_{t}) & 0 & \sin(\theta_{t}) \\
			\sin(\phi_{t})\sin(\theta_{t}) & \cos(\phi_{t}) & -\cos(\theta_{t})\sin(\phi_{t}) \\
			-\cos(\phi_{t})\sin(\theta_{t}) & \sin(\phi_{t}) & \cos(\phi_{t})\cos(\theta_{t})
		\end{bmatrix}
	\end{split}
	\end{equation}

In the top-frame coordinate system $\bm{P_t}$, the position vectors of the three rotors are denoted as $\bm{{}^{P_{t}}r_{i}},i=1,2,3$, and are specifically expressed as: $\bm{^{P_{t}}r_{1}}=[\frac{d}{2}\ \frac{\sqrt{3}}{2}d\ 0]^{T},\bm{^{P_{t}}r_{2}}=[-d\ 0\ 0]^{T},\bm{^{P_{t}}r_{3}}=[\frac{d}{2}\ -\frac{\sqrt{3}}{2}d\ 0]^{T}$.
$d$ is the distance from the geometric center of the top frame to the rotor. The lift generated by the rotor is expressed as $\bm{{}^{P_{t}}F_{i}}=[ \begin{array} {ccc}{0} & {0} & {C_{L}{\omega_{i}}^{2}} \end{array}]^{T}\quad,i=1,2,3$, where $C_{_L}>0$ is the constant-force thrust coefficient, $\omega_{t}$ is the rotational speed of the $i$-th rotor, and the direction of the force vector generated by the rotor acts along the $\bm{P_{tz}}$ axis.

In the top-body coordinate system $\bm{B_{t}}$, the resultant force exerted by the three rotors on the top frame is expressed as:
\begin{equation}
	\bm{{}^{B_{t}}F_{t}}=\sum_{i=1}^{3}\bm{{}^{B_{t}}R_{P_{t}}}\bm{{}^{P_{t}}F_{i}}
   \end{equation}

Due to the counterclockwise rotation of the top rotor, a clockwise torque is generated, denoted as $\bm{^{P_{t}}M_{\tau_{i}}}=[ \begin{array} {ccc}{0} & {0} & {-C_{T}\omega_{i}^{2}} \end{array}]^{T}$, where $C_{T}>0$ represents the rotor drag coefficient. In the top-body coordinate system $\bm{B_{t}}$, the input moment exerted by the rotors on the top frame is expressed as:

\begin{equation}
    \begin{split}
	\bm{{}^{B_{t}}M_{t}} =\sum_{i=1}^{3}\bm{{}^{B_{t}}R_{P_{t}}({}^{P_{t}}r_{i}} \times \bm{{}^{P_{t}}F_{i})}+\sum_{i=1}^{3}\bm{{}^{B_{t}}R_{P_{t}}}\bm{{}^{P_{t}}M_{\tau_{i}}}
    \end{split}   
\end{equation}

The dynamic modeling of the bottom frame follows an analogous formulation: The rotation matrix from the bottom-frame coordinate system $\bm{P_{b}}$ to the bottom-body coordinate system $\bm{B_{b}}$ is denoted as $\bm{{}^{B_b}R_{P_b}}$.
 The position vectors of the three rotors are denoted as $\bm{{}^{P_{b}}r_{j}},j=4,5,6$, and are specifically expressed as: $\bm{^{P_{b}}r_{4}}=[d\ 0 \ 0]^{T},\bm{^{P_{b}}r_{5}}=[-\frac{d}{2}\ \frac{\sqrt{3}}{2}d\ 0]^{T},\bm{^{P_{b}}r_{6}}=[\frac{d}{2}\ -\frac{\sqrt{3}}{2}d\ 0]^{T}$.

In the bottom-body coordinate system $\bm{B_{b}}$, the resultant force exerted by the three rotors on the top frame is expressed as:

\begin{equation}
	\bm{{}^{B_{b}}F_{b}}=\sum_{j=4}^{6}\bm{{}^{B_{b}}R_{P_{b}}}\bm{{}^{P_{b}}F_{j}}
   \end{equation}
   where $\bm{{}^{P_{b}}F_{j}}$ denotes the thrust force generated by the $j$-th rotor in the coordinate system $\bm{P_{b}}$.

The distinction lies in that the rotors rotate clockwise, thus generating a torque opposing the direction of the angular velocity, which is denoted in $\bm{P_{b}}$ as $\bm{{}^{B_{b}}M_{\tau_{j}}}=[0 \ 0 \ C_{T}{\omega_{j}}{}^{2}]^{T}$. In the bottom-body coordinate system $\bm{B_{b}}$, The input moment applied to the bottom frame is expressed as:

\begin{equation}
    \bm{{}^{B_{b}}M_{b}}=\sum_{j=4}^{6}\bm{{}^{B_{b}}R_{P_{b}}({}^{P_{b}}r_{j}} \times \bm{{}^{P_{b}}F_{j})}+\sum_{j=4}^{6}\bm{{}^{B_{b}}R_{P_{b}}}\bm{{}^{P_{b}}M_{\tau_{j}}}
   \end{equation}

Among these, the components of $\bm{{}^{B_t}M_{t}}$ and $\bm{{}^{B_b}M_{b}}$ along the $\bm{X}$-axis and $\bm{Y}$-axis act as disturbances on the top and bottom frames, which are ultimately compensated and eliminated through a feedforward controller. In contrast, the components of $\bm{{}^{B_t}M_{t}}$ and $\bm{{}^{B_b}M_{b}}$ along the $\bm{Z}$-axis directly act on the central body, collectively constituting the yaw moment of the vehicle. For clarity of exposition, the translational dynamics are expressed in the body coordinate system $\bm{B}$, while the rotational dynamics are described in the world inertial coordinate system $\bm{I}$.

Define $\bm{\omega_{_B}}\in\mathbb{R}^3$ as the angular velocity of frame $\bm{B}$ with respect to frame $\bm{I}$, expressed in the body coordinate system $\bm{B}$. The rotational dynamics are then given by:

\begin{equation}
	I_B\bm{\dot{\omega}_B}=-\bm{\omega_B}\times I_B\bm{\omega_B}+\bm{M}
   \end{equation}

where $\bm{I_B}$ is the inertia matrix of the hexacopter, and $\bm{M}$ denotes the total input moment. This moment is decomposed as:

\begin{equation}
	\begin{aligned}
		& \bm{M}=\bm{p_t^B}\times\bm{{}^{B_t}F_t}+\bm{p_b^B}\times\bm{{}^{B_b}F_b}+\bm{{}^{B_t}M_{t}}
	   \begin{bmatrix}
	   0 \\
	   0 \\
	   1
	   \end{bmatrix}+\bm{{}^{B_t}M_{b}}
	   \begin{bmatrix}
	   0 \\
	   0 \\
	   1
	   \end{bmatrix} \\
		& 
	   \end{aligned}
   \end{equation}

The translational dynamics can be expressed in the inertial frame $\bm{I}$ using the standard Newton-Euler formulation as follows:

\begin{equation}
	m\bm{\ddot{p}}=m
   \begin{bmatrix}
   0 \\
   0 \\
   -\bm{g}
   \end{bmatrix}+\bm{{}^WR_B}(\bm{{}^{B_{t}}F_{t}}+\bm{{}^{B_{b}}F_{b}})
   \end{equation}

   where $\bm{\ddot{p}}= \begin{bmatrix} \ddot{x}_m \ \ddot{y}_m \ \ddot{z}_m \end{bmatrix}^T$ represents the acceleration of $\bm{B}$ in $\bm{I}$, and ${}^{W}{R_B}$ is the rotation matrix from the body coordinate system $\bm{B}$ to the world inertial coordinate system $\bm{I}$.

\section{Controller Design}

To address the dynamic characteristics of the fully-actuated hexacopter, this paper proposes a novel hierarchical control system architecture for fully-actuated hexacopter. The control framework consists of two layers: a body controller and top/bottom controllers. These layers collectively achieve six-degree-of-freedom motion control through information interaction and moment distribution.

\subsection{Body Controller Design}

The body controller serves as the core control unit, which is divided into four components: a body position controller, a body velocity controller, a body attitude controller, and a body angular velocity controller. A dual-loop cascaded PID control strategy is employed in this architecture.

The body position controller and the body attitude controller receive commands for the desired three-axis position $[x_{md},y_{md},z_{md}]$ and the desired three-axis attitude $[\phi_{md},\theta_{md},\psi_{md}]$ from a high-level planner, respectively. They output the desired three-axis velocity $[\dot{x}_{md},\dot{y}_{md},\dot{z}_{md}]$ and the desired three-axis angular velocity $[\dot{\phi}_{md},\dot{\theta}_{md},\dot{\psi}_{md}]$. Meanwhile, the actual three-axis position $[x_{m},y_{m},z_{m}]$ and velocity $[\dot{x}_{m},\dot{y}_{m},\dot{z}_{m}]$ of the vehicle are measured utilizing a UWB system, while the actual three-axis attitude $[\phi_{m},\theta_{m},\psi_{m}]$ and angular velocity $[\dot{\phi}_{m},\dot{\theta}_{m},\dot{\psi}_{m}]$ are measured via a BMI088 sensor. This enables the implementation of a cascaded PID control strategy for the position-velocity loop and the attitude-angular velocity loop, respectively. The control laws are as follows:

\begin{equation}
	u_{m}=K_{p,\tau}^{a}(\tau_{md}-\tau_{m})+K_{d,\tau}^{a}(\dot{\tau}_{md}-\dot{\tau}_{m})+K_{i,\tau}^{a}\int(\tau_{md}-\tau_{m})
\end{equation}

where $\tau_{m}$ represents the actual vehicle state, $\tau_{m}\triangleq\{x_{m},y_{m},z_{m},\dot{x}_{m},\dot{y}_{m},\dot{z}_{m},\phi_{m},\theta_{m},\psi_{m},\dot{\phi}_{m},\dot\theta_{m},\dot{\psi}_{m}\}$

\subsection{Top/Bottom Controller Design}

The top and bottom  controllers comprise four components: a top attitude controller, a top angular velocity controller, a bottom attitude controller, and a bottom angular velocity controller. As derived from the earlier modeling, the two thrust vectors generated by the top and bottom frames can be combined in two distinct methods. This enables independent control of the vehicle's translation and attitude. Consequently, the desired roll and pitch angles for the top-and bottom attitude controllers can be expressed as:

\begin{subequations}
	\begin{align}
		\phi_{td}& = \ddot{\phi}_{md} + \ddot{y}_{md},  \theta_{td} = \ddot{\theta}_{md} + \ddot{x}_{md} \\
		\phi_{bd}& = -\ddot{\phi}_{md} + \ddot{y}_{md}, \theta_{bd} = -\ddot{\theta}_{md} + \ddot{x}_{md}
		\end{align} 
\end{subequations}

The Top/Bottom Attitude Controller outputs the desired roll and pitch angular velocities $\dot{\phi}_{td},\dot{\theta}_{td},\dot{\phi}_{bd},\dot{\theta}_{bd}$, which serve as inputs to the Top/Bottom Angular Rate Controller. Through the Top/Bottom MPU6050 sensors, the real-time roll and pitch angles $[\phi_t,\theta_t,\phi_b,\theta_b]$ and angular velocities $[\dot{\phi}_t,\dot{\theta}_t,\dot{\phi}_b,\dot{\theta}_b]$ are measured. This enables the implementation of cascade PID control for the Top/Bottom attitude-angular velocity loops. The control law is expressed as:
\begin{equation}
	u=K_{p,\tau}^a(\tau_d-\tau)+K_{d,\tau}^a(\dot{\tau}_d-\dot{\tau})+K_{i,\tau}^a\int(\tau_d-\tau)
\end{equation}

where $\tau\triangleq\{\phi_t,\theta_t,\phi_b,\theta_b,\dot{\phi}_t,\dot{\theta}_t,\dot{\phi}_b,\dot{\theta}_b\}$, and $u$ denotes the control output corresponding to different values of $\tau$.

\subsection{Thrust Allocation Matrix}

In this structural design, rotation of the top-frame rotors generates clockwise torque, while rotation of the bottom-frame rotors produces counterclockwise torque. The overall yaw motion is achieved through differential torque between the top and bottom frames. When a desired yaw angular velocity $\dot\psi_{md}$ (positive in the counterclockwise direction) is commanded, the body angular velocity controller outputs a positive angular acceleration $\ddot{\psi}_{md}$. To realize this motion, the input to the top-frame motors is reduced by $\ddot{\psi}_{md}$, while the input to the bottom-frame motors is increased by $\ddot{\psi}_{md}$, thereby generating a net yaw torque for positive directional rotation.

The computation of the yaw angle for the body, top frame, and bottom frame is fully integrated and executed by the body controller, ensuring unified attitude coordination across all structural components. Subsequently, the control outputs from the top angular velocity controller, bottom angular velocity controller, and altitude controller are synthesized and dynamically distributed to the six motors. This distribution follows specific thrust allocation relationships, enabling the vehicle to achieve highly agile and fully-actuated motion in all degrees of freedom. Based on the unique configuration design of the fully-actuated hexacopter platform, the following motor thrust allocation matrix is systematically derived:
\begin{subequations}
	\begin{align}
		\begin{bmatrix}
			u_1 \\
			u_2 \\
			u_3
		\end{bmatrix} &= k
		\begin{bmatrix}
			1 & -\frac{1}{2} & 1 & -1 \\
			1 & 1 & 0 & -1 \\
			1 & -\frac{1}{2} & -1 & -1
		\end{bmatrix}
		\begin{bmatrix}
			\ddot{z}_{md} \\
			\ddot{\theta}_{td} \\
			\ddot{\phi}_{td} \\
			\ddot{\psi}_{md}
		\end{bmatrix} \\
		\begin{bmatrix}
			u_4 \\
			u_5 \\
			u_6
		\end{bmatrix} &= k
		\begin{bmatrix}
			1 & -1 & 0 & \;\;\;1 \\
			1 & \ \ \, \frac{1}{2} & 1 & \ \, \, 1 \\
			1 & \ \ \, \frac{1}{2} & -1 & \ \, \, 1
		\end{bmatrix}
		\begin{bmatrix}
			\ddot{z}_{md} \\
			\ddot{\theta}_{bd} \\
			\ddot{\phi}_{bd} \\
			\ddot{\psi}_{md}
		\end{bmatrix}
	\end{align}
	\end{subequations}
where $u_1,u_2,u_3,u_4,u_5,u_6$ denotes the PWM output and $k$ represents the output coefficient.

\section{Simulation Studies}

In this simulation experiment, the desired trajectory for the fully-actuated hexacopter is a circular path with a radius of 1 m and a height of 1.5 m. The center of the trajectory is located at (2.3, 3.5, 1.5) m. 

\begin{figure}[htbp]
    \centering
    \subfigure[]{
        \includegraphics[width=0.4\textwidth]{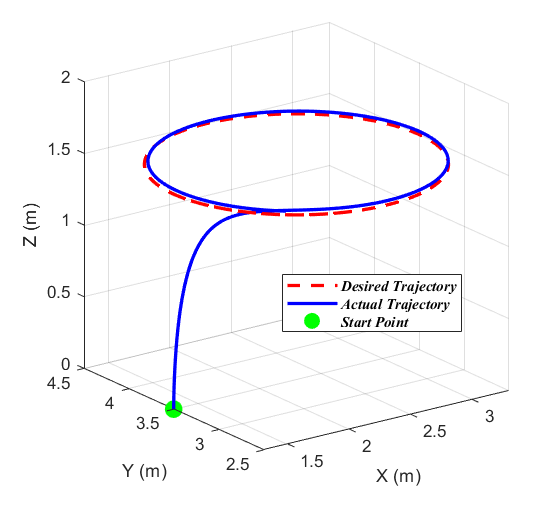}
        \label{1} 
    }
    \vspace{-0.5em} 
 
    \subfigure[]{
        \includegraphics[width=0.4\textwidth]{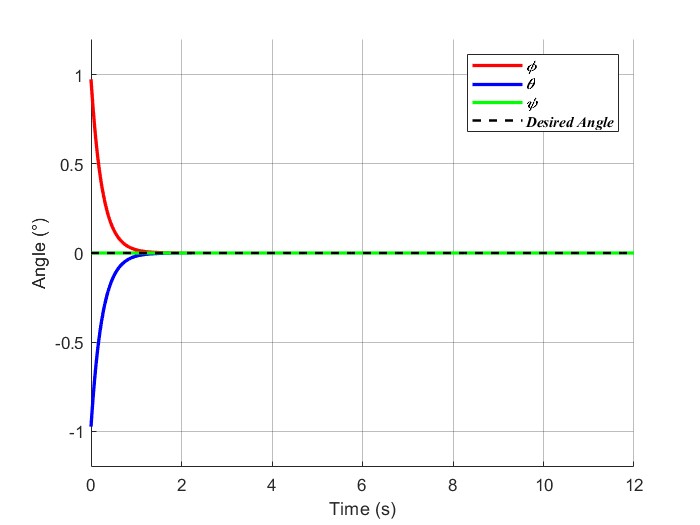}
        \label{2} 
    }
    \vspace{0em} 
    \caption{Flight trajectory simulation test results. (a) the fully-actuated hexacopter flight trajectory, with the desired trajectory being a circle with a diameter of 1 meters. (b) Time response of the attitude.} 
    \label{Flight trajectory} 
\end{figure}
At the start of the experiment, the fully-actuated hexacopter takes off from the initial position $\mathbf{p}_{\mathrm{g}}(0)= \begin{bmatrix} 1.3,3.5,0 \end{bmatrix}^{\mathrm{T}}$ m with an initial attitude of $\mathbf{\Theta}(0)= \begin{bmatrix} 1,-1,0 \end{bmatrix}^\mathrm{T}$ deg. Other relevant simulation parameters are set as follows:
$$\begin{aligned}
 & \mathbf{p}_{\mathrm{gd}}=\left[2.3-\cos(\frac{\pi}{5}t),3.5-\sin(\frac{\pi}{5}t),1.5\right]^\mathrm{T}\mathrm{m}, \\
 & \mathbf{g}=9.8\mathrm{kg}\cdot\mathrm{m}^{-2},m=0.8\mathrm{kg}, \\
 & J=\mathrm{diag}(0.0120,0.0115,0.0024)\mathrm{kg}\cdot\mathrm{m}^2
\end{aligned}$$

As shown in Fig \ref{Flight trajectory}, the actual flight trajectory of the fully-actuated hexacopter and the curve of its attitude variation over time are presented. The results indicate that the hexacopter is capable of following the circular path while maintaining a consistent attitude, thereby validating the synchronization control of position and attitude under the full-actuation controller.

\section{Conclusion}

This paper investigates the design, modeling, and control of fully-actuated hexacopter. We innovatively propose a dual-frame passive tilting structure for fully-actuated hexacopter and complete its dynamic modeling and controller design. This structure achieves independent control of translational motion and attitude through frame tilting and rotor differential speed. Experiments validate the fully-actuated motion capability of the fully-actuated hexacopter platform and the effectiveness of control strategy. 

In future work, a physical experimental platform for the fully-actuated hexacopter will be constructed to validate its full-actuation characteristics through systematic experiments. Subsequently, a robust wind-disturbance rejection controller will be designed to enhance the adaptability and stability of fully-actuated hexacopter in complex wind environments. Furthermore, the energy consumption characteristics will be analytically evaluated to provide insights for optimizing energy efficiency and extending operational endurance in practical applications.


\begin{thebibliography}{00}


\bibitem{b1} J. Chen, T. Liu and S. Shen, ``Tracking a moving target in cluttered environments using a quadrotor,'' 2016 IEEE/RSJ International Conference on Intelligent Robots and Systems (IROS), Daejeon, Korea (South), 2016, pp. 446-453.
\bibitem{b2} Z. Wang, B. Liu, X. Cui, Y. Li, Y. Yang and Z. Lyu, ``The Application of Micro Coaxial Rotorcraft in Warfare: An Overview, Key Technologies, and Warfare Scenarios,'' in IEEE Access, vol. 10, pp. 40358-40366, 2022.
\bibitem{b3} X. Zhao and H. Zhou, ``Distributed Collaborative Control Technology in Multiple Unmanned Aerial Vehicle Operations'' 2023 5th International Conference on Robotics, Intelligent Control and Artificial Intelligence (RICAI), Hangzhou, China, 2023, pp. 393-396.
\bibitem{b4} C. Kerr, R. Jaradat and N. U. Ibne Hossain, ``Battlefield Mapping by an Unmanned Aerial Vehicle Swarm: Applied Systems Engineering Processes and Architectural Considerations From System of Systems'' in IEEE Access, vol. 8, pp. 20892-20903, 2020.
\bibitem{b5} A. Alsawy, A. Hicks, D. Moss and S. Mckeever, ``An Image Processing Based Classifier to Support Safe Dropping for Delivery-by-Drone'' 2022 IEEE 5th International Conference on Image Processing Applications and Systems (IPAS), Genova, Italy, 2022, pp. 1-5.
\bibitem{b6} J. Deng, Y. He, B. Gao and P. Liu, ``A Drone Delivery Problem Arising in the Meal Delivery based on VNS Algorithm,'' 2024 4th International Conference on Control Theory and Applications (ICoCTA), Hangzhou, China, 2024, pp. 123-127.
\bibitem{b7} R. Rashad, F. Califano and S. Stramigioli, ``Port-Hamiltonian Passivity-Based Control on SE(3) of a Fully Actuated UAV for Aerial Physical Interaction Near-Hovering,'' in IEEE Robotics and Automation Letters, vol. 4, no. 4, pp. 4378-4385, Oct. 2019.
\bibitem{b8} D. Kotarski, P. Piljek, H. Brezak and J. Kasać, ``Design of a fully actuated passively tilted multirotor UAV with decoupling control system,'' 2017 8th International Conference on Mechanical and Aerospace Engineering (ICMAE), Prague, Czech Republic, 2017, pp. 385-390.
\bibitem{b9} Z. Li, H. Wang, X. Zhang, Y. Liu, X. Zhang and Y. Zhuang, ``Nonlinear Control of a Fully-Actuated Hexacopter with a Cable-Suspended Load,'' 2024 14th Asian Control Conference (ASCC), Dalian, China, 2024, pp. 2340-2345.
\bibitem{b10} A. Flores and G. Flores, ``Fully actuated Hexa-rotor UAV: Design, construction, and control. Simulation and experimental validation,'' 2022 International Conference on Unmanned Aircraft Systems (ICUAS), Dubrovnik, Croatia, 2022, pp. 1497-1503.
\bibitem{b11} P. Abbaraju, X. Ma, G. Jiang, M. Rastgaar and R. M. Voyles, ``Aerodynamic Modeling of Fully-Actuated Multirotor UAVs with Nonparallel Actuators,'' 2021 IEEE/RSJ International Conference on Intelligent Robots and Systems (IROS), Prague, Czech Republic, 2021, pp. 9639-9645.
\bibitem{b12} Y. Aboudorra, A. Saini and A. Franchi, ``Gain Scheduling Position Control for Fully-Actuated Morphing Multi-Rotor UAVs,'' 2024 International Conference on Unmanned Aircraft Systems (ICUAS), Chania - Crete, Greece, 2024, pp. 15-22.
\bibitem{b13} N. Staub, D. Bicego, Q. Sablé, V. Arellano, S. Mishra and A. Franchi, ``Towards a Flying Assistant Paradigm: the OTHex,'' 2018 IEEE International Conference on Robotics and Automation (ICRA), Brisbane, QLD, Australia, 2018, pp. 6997-7002.
\bibitem{b14} Z. Zhu, J. Yu, Y. Lin and Y. Zhang, ``Design, Modeling and Adaptive Robust Control of a Spatial Symmetric Omni-Directional Aerial Vehicle,'' in IEEE Robotics and Automation Letters, vol. 9, no. 6, pp. 5958-5965, June 2024.
\bibitem{b15} S. Park, J. Her, J. Kim and D. Lee, ``Design, modeling and control of omni-directional aerial robot,'' 2016 IEEE/RSJ International Conference on Intelligent Robots and Systems (IROS), Daejeon, Korea (South), 2016, pp. 1570-1575.
\bibitem{b16} H. Lee, S. Cheng, Z. Wu, J. Lim, R. Siegwart and N. Hovakimyan, ``Geometric Tracking Control of Omnidirectional Multirotors for Aggressive Maneuvers,'' in IEEE Robotics and Automation Letters, vol. 10, no. 2, pp. 1130-1137, Feb. 2025.
\bibitem{b17} M. Ryll, D. Bicego, M. Giurato, M. Lovera and A. Franchi, ``FAST-Hex—A Morphing Hexarotor: Design, Mechanical Implementation, Control and Experimental Validation,'' in IEEE/ASME Transactions on Mechatronics, vol. 27, no. 3, pp. 1244-1255, June 2022.
\bibitem{b18} M. Kamel et al., ``The Voliro Omniorientational Hexacopter: An Agile and Maneuverable Tiltable-Rotor Aerial Vehicle,'' in \emph{IEEE Robotics \& Automation Magazine}, vol. 25, no. 4, pp. 34--44, Dec. 2018.
\bibitem{b19} B. Li, L. Ma, D. Huang, X. Sun and T. Hou, ``Designing, Modeling, and Control Allocation Strategy of a Novel Omnidirectional Aerial Robot,'' 2019 IEEE Conference on Control Technology and Applications (CCTA), Hong Kong, China, 2019, pp. 762-767.
\bibitem{b20} Z. Qin, J. Wei, M. Cao, B. Chen, K. Li and K. Liu, ``Design and Flight Control of a Novel Tilt-Rotor Octocopter Using Passive Hinges,'' in IEEE Robotics and Automation Letters, vol. 9, no. 1, pp. 199-206, Jan. 2024.


\end{thebibliography}
\end{document}